\documentclass[11pt,a4paper]{article}
\usepackage[hyperref]{acl2019}
\usepackage{times}
\usepackage{latexsym}
\usepackage{url}

\usepackage{array}
\usepackage{float}
\usepackage{CJK}
\usepackage{amsfonts}
\usepackage{amsmath}
\usepackage{subfigure}
\usepackage{graphicx}
\usepackage[export]{adjustbox} 
\graphicspath{ {./graph/} }
\usepackage{mathtools}
\usepackage{booktabs}
\usepackage{makecell}
\usepackage{multirow}
\usepackage{xcolor}
\usepackage{tabularx}

\usepackage[ruled,vlined]{algorithm2e}
\include{pythonlisting}
\usepackage[noend]{algpseudocode}

\aclfinalcopy 
\def\aclpaperid{602} 

\title{Memory Consolidation for Contextual Spoken Language Understanding with Dialogue Logistic Inference}

\author{
	He Bai$^{1,2}$, 
	Yu Zhou$^{1,2}$,
    Jiajun Zhang$^{1,2}$ and
	Chengqing Zong$^{1,2,3}$
	\\ 
	$^1$ National Laboratory of Pattern Recognition, Institute of Automation, CAS, Beijing, China\\
	$^2$ University of Chinese Academy of Sciences, Beijing, China\\
	$^3$ CAS Center for Excellence in Brain Science and Intelligence Technology\\
	\{he.bai, yzhou, jjzhang, cqzong\}@nlpr.ia.ac.cn
}

\date{}

\begin{document}
\maketitle
\begin{abstract}
Dialogue contexts are proven helpful in the spoken language understanding (SLU) system and they are typically encoded with explicit memory representations. However, most of the previous models learn the context memory with only one objective to maximizing the SLU performance, leaving the context memory under-exploited. In this paper, we propose a new dialogue logistic inference (DLI) task to consolidate the context memory jointly with SLU in the multi-task framework. DLI is defined as sorting a shuffled dialogue session into its original logical order and shares the same memory encoder and retrieval mechanism as the SLU model. Our experimental results show that various popular contextual SLU models can benefit from our approach, and improvements are quite impressive, especially in slot filling.
\end{abstract}
\section{Introduction}
\begin{figure}[ht]
 \centering
 \includegraphics[width=0.5\textwidth]{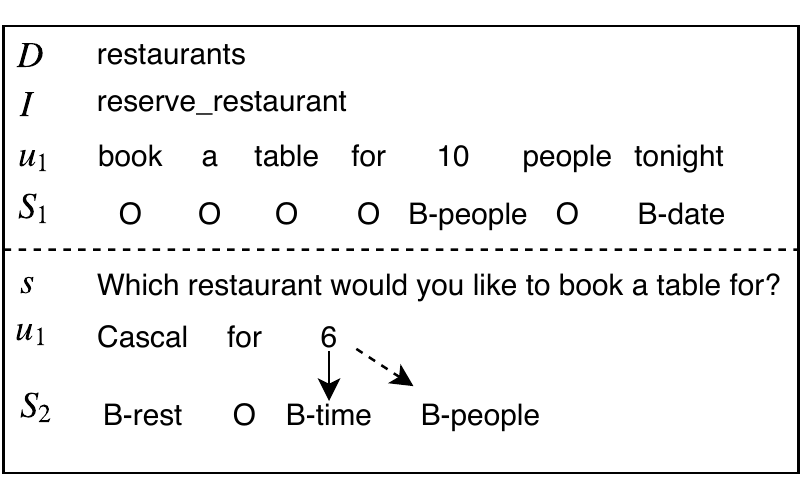}
 \caption{Example semantic frames of utterances $u_1$ and $u_2$ with domain (D), intent (I) and semantic slots in IOB format ($S_1$, $S_2$).}
 \label{fig:semantic_form}
\end{figure}

Spoken language understanding (SLU) is a key technique in today's conversational systems such as Apple Siri, Amazon Alexa, and Microsoft Cortana. A typical pipeline of SLU includes domain classification, intent detection, and slot filling\cite{tur2011spoken}, to parse user utterances into semantic frames. Example semantic frames \cite{coling_tutorial} are shown in Figure \ref{fig:semantic_form} for a restaurant reservation.

Traditionally, domain classification and intent detection are treated as classification tasks with popular classifiers such as support vector machine and deep neural network \cite{haffner2003optimizing,sarikaya2011deep}. They can also be combined into one task if there are not many intents of each domain\cite{bai2018source}. Slot filling task is usually treated as a sequence labeling task. Popular approaches for slot filling include conditional random fields (CRF) and recurrent neural network (RNN) \cite{raymond2007generative,yao2014spoken}. Considering that pipeline approaches usually suffer from error propagation, the joint model for slot filling and intent detection has been proposed to improve sentence-level semantics via mutual enhancement between two tasks \cite{xu2013convolutional,hakkani2016multi,zhang2016joint,goo_slot-gated_2018}, which is a direction we follow.
\begin{figure*}[ht!]
  \centering
  \includegraphics[width=0.8\textwidth]{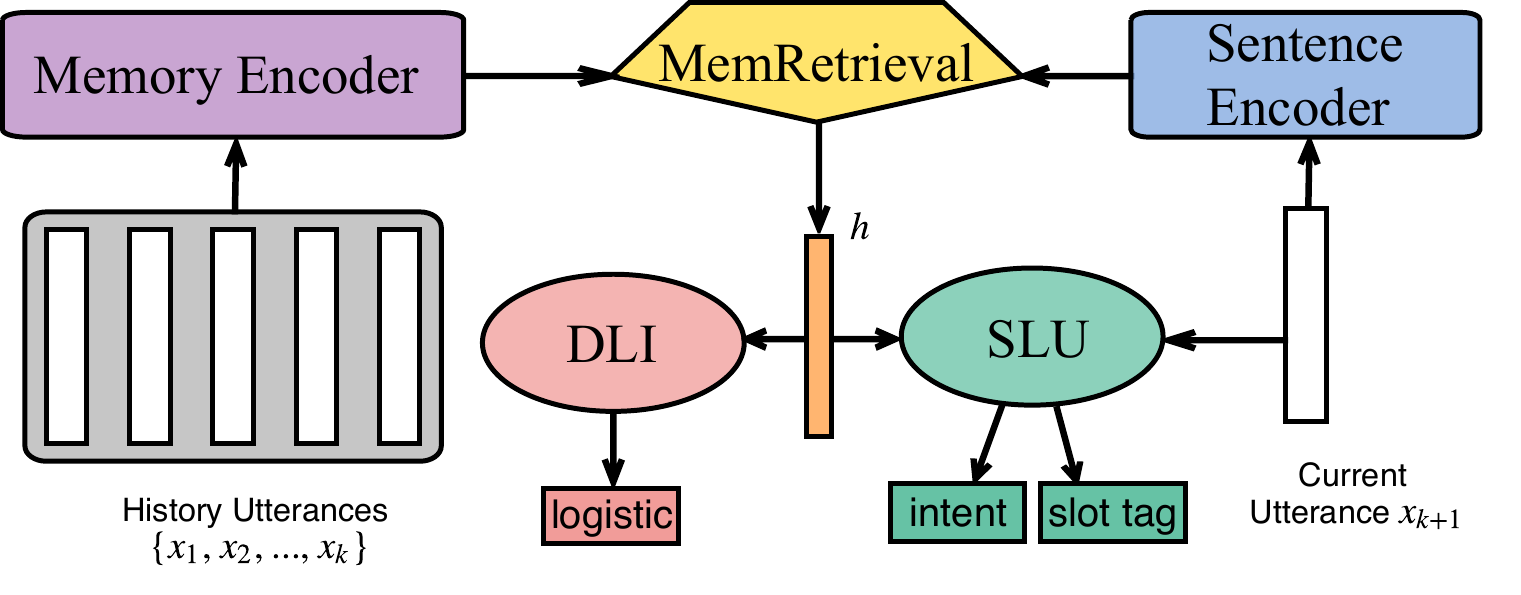}
  \caption{
  Architecture of our proposed contextual SLU with memory consolidation.
  }
  \label{fig:overall}
\end{figure*}
To create a more effective SLU system, the contextual information has been shown useful \cite{bhargava_easy_2013,xu_contextual_2014}, as natural language utterances are often ambiguous. For example, the number 6 of utterance $u_2$ in Figure \ref{fig:semantic_form} may refer to either B-time or B-people without considering the context. Popular contextual SLU models \cite{chenEndtoEndMemoryNetworks2016,bapnaSequentialDialogueContext2017} exploit the dialogue history with the memory network \cite{weston_memory_2015}, which covers all three main stages of memory process: encoding (write), storage (save) and retrieval (read) \cite{baddeley1976psychology}. With such a memory mechanism, SLU model can retrieve context knowledge to reduce the ambiguity of the current utterance, contributing to a stronger SLU model. However, the memory consolidation, a well-recognized operation for maintaining and updating memory in cognitive psychology \cite{sternberg2016cognitive}, is underestimated in previous models. They update memory with only one objective to maximizing the SLU performance, leaving the context memory under-exploited.

In this paper, we propose a multi-task learning approach for multi-turn SLU by consolidating context memory with an additional task: dialogue logistic inference (DLI), defined as sorting a shuffled dialogue session into its original logical order. DLI can be trained with contextual SLU jointly if utterances are sorted one by one: selecting the right utterance from remaining candidates based on previously sorted context. In other words, given a response and its context, the DLI task requires our model to infer whether the response is the right one that matches the dialogue context, similar to the next sentence prediction task \cite{logeswaran_efficient_2018}. We conduct our experiments on the public multi-turn dialogue dataset KVRET \cite{eric_key-value_2017}, with two popular memory based contextual SLU models. According to our experimental results, noticeable improvements are observed, especially on slot filling.

\section{Model Architecture}

This section first explains the memory mechanism for contextual SLU, including memory encoding and memory retrieval. Then we introduce the SLU tagger with context knowledge, the definition of DLI and how to optimize the SLU and DLI jointly. The overall model architecture is illustrated in Figure \ref{fig:overall}.
\subsection*{Memory Encoding}
To represent and store dialogue history $\{x_1, x_2, ... x_k\}$, we first encode them into memory embedding $M=\{m_1, m_2, ... m_k\}$ with a BiGRU \cite{chung2014empirical} layer and then encode the current utterance $x_{k+1}$ into sentence embedding $c$ with another BiGRU:
\begin{equation}\label{equ:encoder}
 m_i=\text{BiGRU}_m(x_i) \quad 
 c=\text{BiGRU}_c(x_{k+1})
\end{equation}
\subsection*{Memory Retrieval}

Memory retrieval refers to formulating contextual knowledge of the user's current utterance $x_{k+1}$ by recalling dialogue history. 
There are two popular memory retrieval methods:

The \textbf{attention based} \cite{chenEndtoEndMemoryNetworks2016} method first calculates the attention distribution of $c$ over memories $M$ by taking the inner product followed by a softmax function. Then the context can be represented with a weighted sum over $M$ by the attention distribution:
\begin{equation}\label{equ:attention}
 p_i=\text{softmax}(c^Tm_i) \quad
 m_{ws} =\sum_ip_im_i
\end{equation}
where $p_i$ is the attention weight of $m_i$. In \citeauthor{chenEndtoEndMemoryNetworks2016}, they sum $m_{ws}$ with utterance embedding $c$, then multiplied with a weight matrix $W_o$ to generate an output knowledge encoding vector $h$:
\begin{align}\label{equ:add_h}
 h&=W_o(c+m_{ws})
\end{align}
The \textbf{sequential encoder based} \cite{bapnaSequentialDialogueContext2017} method shows another way to calculate $h$: 
\begin{align}\label{equ:SDEN}
 g_i&=\text{sigmoid}(\text{FF}([c\,;m_i]))\\
 h&=\text{BiGRU}_g([g_1,g_2,...,g_k])
\end{align}
where the function $\text{FF}()$ is a fully connected forward layer. 

\subsection*{Contextual SLU}
Following \citeauthor{bapnaSequentialDialogueContext2017}, our SLU model is a stacked BiRNN: a BiGRU layer followed by a BiLSTM layer. However, \citeauthor{bapnaSequentialDialogueContext2017} only initializes the BiLSTM layer's hidden state with $h$, resulting in the low participation of context knowledge. In this work, we feed $h$ to the second layer in every time step:
\begin{align}\label{equ:sden_h}
O_1 &= \text{BiGRU}_1(x_{k+1})\\
O_2& = \text{BiLSTM}_2([O_1; h])
\end{align}
where $O_1={\{o_1^1, ..., o_1^m\}}$ is the first layer's output and $m$ is the length of $x_{k+1}$. The second layer encodes $\{[o_1^1\,;h], ..., [o_1^m\,;h]\}$ into the final state \(s_2 = [\overrightarrow{s_2}\,;\overleftarrow{s_2}]\) and outputs $O_2 = {\{o_2^1, ..., o_2^m\}}$, which can be used in the following intent detection layer and slot tagger layer respectively.
\begin{equation}\label{equ:output_layer}
 P^{i}=\text{softmax}(Us_2)\quad
 P^{s}_t=\text{softmax}(Vo_2^t)
\end{equation}
where $U$ and $V$ are weight matrices of output layers and $t$ is the index of each word in utterance $x_{k+1}$.

\subsection*{Dialogue Logistic Inference}
As described above, the memory mechanism holds the key to contextual SLU. However, context memory learned only with SLU objective is under-exploited. Thus, we design a dialogue logistic inference (DLI) task that can consolidate the context memory by sharing encoding and retrieval components with SLU. DLI is introduced below:

Given a dialogue session $X=\{x_1, x_2, ... x_n\}$, where $x_i$ is the $ith$ sentence in this conversation, we can shuffle $X$ into a random order set $X'$. It is not hard for human to restore $X'$ to $X$ by determining which is the first sentence then the second and so on. This is the basic idea of DLI: choosing the right response given a context and all candidates. For each integer $j$ in range $k+1$ to $n$, training data of DLI can be labelled automatically by:
\begin{align}\label{equ:DLI}
P(x_j|x_1,...,x_k)= \left\{
\begin{array}{lrc}
1 &&{j = k+1}\\
0 &&{j \neq k+1}
\end{array} 
\right.
\end{align}
where $k+1$ is the index of the current utterance. In this work, we calculate the above probability with a 2-dimension softmax layer:
\begin{align}
 P(x_j|x_1,...,x_{k}) = \text{softmax}(W_{d}h)
\end{align}
where $W_{d}$ is a weight matrix for dimension transformation.

\subsection*{Joint Optimization}
As we depict in Figure \ref{fig:overall}, we train DLI and SLU jointly in order to benefit the memory encoder and memory retrieval components. Loss functions of SLU and DLI are as follows.
\begin{align}\label{equ:loss_i}
\begin{split}
L_{SLU}& = \text{log}(p(y^I|x_1,...,x_{k+1}))\\ &+\sum_t\text{log}(p(y^S_t|x_1,...,x_{k+1}))
\end{split}\\
L_{DLI}& = \sum_{x_j}\text{log}(p(y^D|x_j, x_1,...,x_k))
\end{align}
where $x_j$ is a candidate of the current response, $y^I$, $y_t^S$ and $y^D$ are training targets of intent, slot and DLI respectively. Finally, the overall multi-task loss function is formulated as 
\begin{align}\label{equ:loss_all}
L = (1-\lambda)L_{SLU} + \lambda L_{DLI}
\end{align}
where $\lambda$ is a hyper parameter. 
\section{Experiments}

In this section, we first introduce datasets we used, then present our experimental setup and results on these datasets.

\subsection{Datasets}
KVRET \cite{eric_key-value_2017} is a multi-turn task-oriented dialogue dataset for an in-car assistant. This dataset was collected with the Wizard-of-Oz scheme \cite{wen_network-based_2017} and consists of 3,031 multi-turn dialogues in three distinct domains, and each domain has only one intent, including calendar scheduling, weather information retrieval, and point-of-interest navigation.

However, all dialogue sessions of KVRET are single domain. Following \citeauthor{bapnaSequentialDialogueContext2017}, we further construct a multi-domain dataset KVRET* by randomly selecting two dialogue sessions with different domain from KVRET and recombining them into one conversation. The recombining probability is set to 50\%. Detailed information about these two datasets is shown in Table \ref{table:datasets}.
\begin{table}[t]
\begin{tabular}{l|c|c|c|c}
\hline
\textbf{Datasets} &Train&Dev&Test&Avg.turns \\
\hline
KVRET &2425&302&304&5.25\\
\hline
KVRET* &1830&224&226&6.88\\
\hline
\end{tabular}
\caption{\label{table:datasets}Detailed information of KVRET and KVRET* datasets, including train/dev/test size and average turns per conversation.} 
\end{table}

\begin{table*}[ht]
\centering
\begin{tabular}{|l|*{9}{l|}}
\hline
\multirow{3}{*}{\textbf{Models}} & \multirow{3}{*}{\textbf{DLI}} &\multicolumn{4}{c}{\textbf{KVRET}} & \multicolumn{4}{|c|}{\textbf{KVRET*}}\\
\cline{3-10}
& &\multicolumn{3}{c|}{\textbf{Slot}} & \textbf{Intent}&\multicolumn{3}{c|}{\textbf{Slot}} & \textbf{Intent}\\
\cline{3-10}
 & &P&R&F1&Acc.&P&R&F1&Acc.\\ 
\hline

NoMem &No&54.8 &80.0&56.7&93.4
&48.9 &81.0&54.7&93.8\\
\hline

\multirow{2}{*}{MemNet} &No &75.8 &81.1&75.8&93.9
&73.1&81.8&74.5&92.8\\
\cline{2-10}
 &Yes&76.0 & 82.3 & \textbf{77.4}(\textbf{+1.6}) & 93.9(+0)
 &75.8 & 81.3 & 76.3(\textbf{+1.8})& \textbf{93.8}(\textbf{+1.0})\\
\hline

\multirow{2}{*}{SDEN } &No &70.5&80.9&70.1&93.6
&56.9&81.3&59.4&93.0\\
\cline{2-10}
 &Yes&64.9 & 80.9 & 70.8 (\textbf{+0.7})& 93.8(\textbf{+0.2})
 &56.5&81.4&60.2(\textbf{+0.8})&93.5\textbf{(+0.5)}\\
\hline

\multirow{2}{*}{$\text{SDEN}^\dagger$ } &No &71.9&82.2&74.0&93.7&
72.7&80.8&74.9&93.2 \\
\cline{2-10}
 &Yes&75.2&81.4&76.6(\textbf{+2.6})&\textbf{94.3}(\textbf{+0.6})
 &78.0&81.4&\textbf{78.3}(\textbf{+3.4})&93.2(+0) \\
\hline

\end{tabular}
\caption{\label{table:results} SLU results on original KVRET and multi-domain KVRET*, including accuracy of intent detection and average precision, recall and F1 score of slot filling. } 
\end{table*}
\begin{figure*}[th]
\centering
 \subfigure[]{\includegraphics[width=0.47\textwidth]{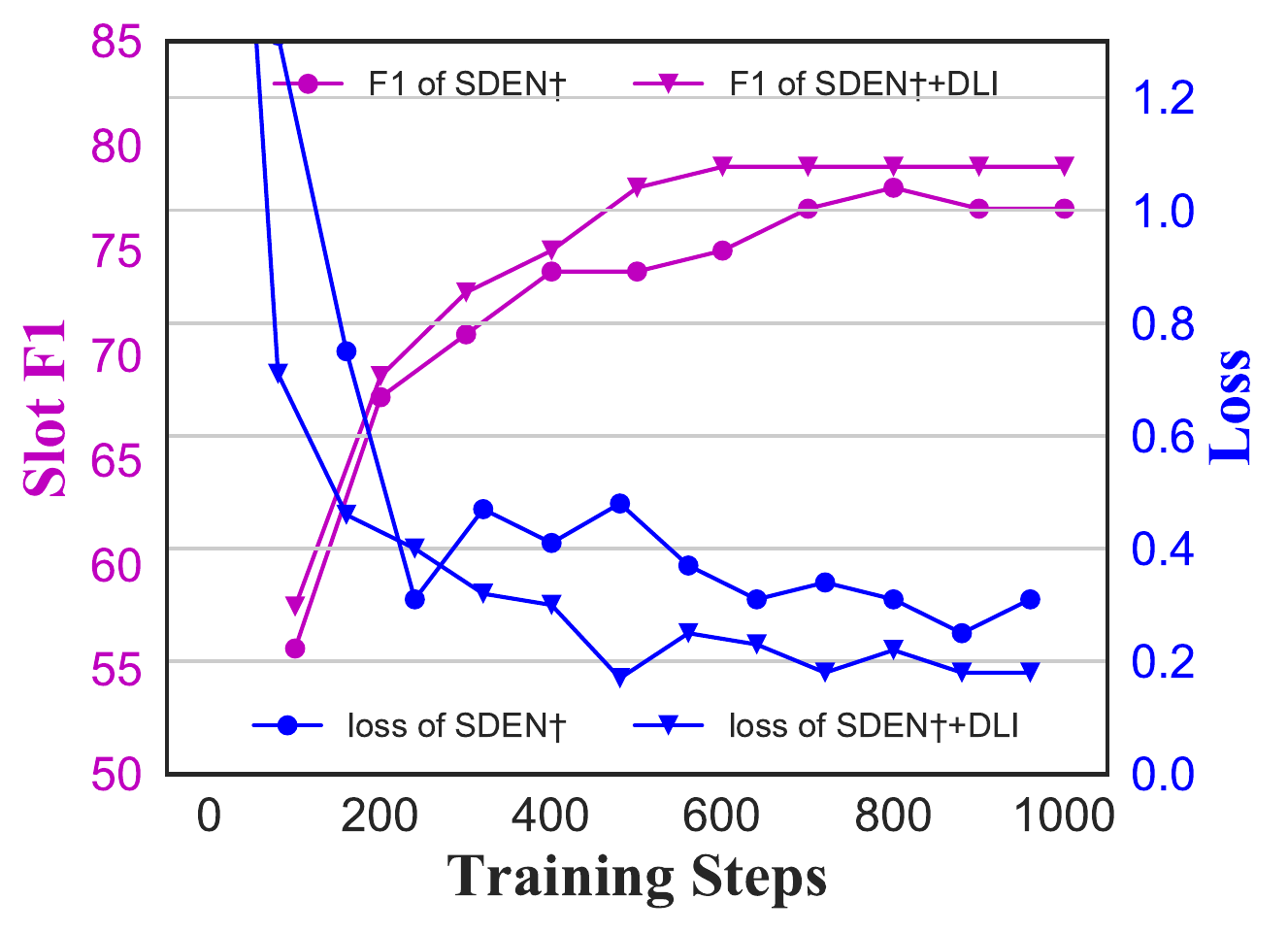}\label{fig:loss on validation set}}
 \subfigure[] 
 {\includegraphics[ width=0.47\textwidth]{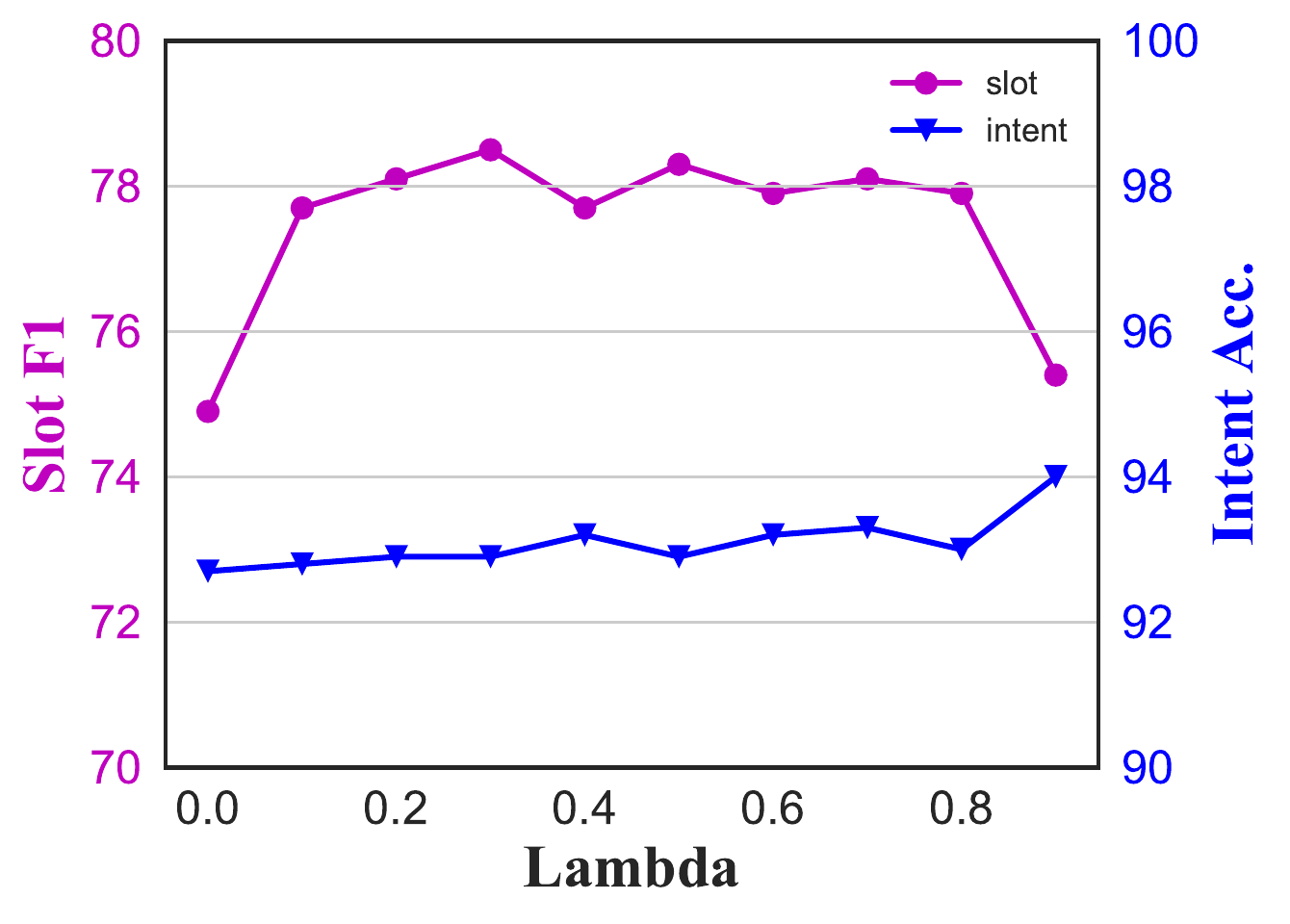}\label{fig: lambda}}
 \caption{(a) Validation loss and slot F1 score of $\text{SDEN}^\dagger$ during training. (b) Slot F1 score and intent accuracy of $\text{SDEN}^\dagger$ with different lambda.}
 \label{fig:curves}
\end{figure*}
\subsection{Experimental Setup}
We conduct extensive experiments on intent detection and slot filling with datasets described above. The domain classification is skipped because intents and domains are the same for KVRET.

For training model, our training batch size is 64, and we train all models with Adam optimizer with default parameters \cite{kingma2014adam}. For each model, we conduct training up to 30 epochs with five epochs' early stop on validation loss. The word embedding size is 100, and the hidden size of all RNN layer is 64. The $\lambda$ is set to be 0.3. The dropout rate is set to be 0.3 to avoid over-fitting.

\subsection{Results}
The following methods are investigated and their results are shown in Table \ref{table:results}:

\textbf{NoMem:} A single-turn SLU model 
without memory mechanism. 

\textbf{MemNet:} The model described in \citeauthor{chenEndtoEndMemoryNetworks2016} , with attention based memory retrieval.




\textbf{SDEN:} The model described in \citeauthor{bapnaSequentialDialogueContext2017} , with sequential encoder based memory retrieval.


\textbf{$\text{SDEN}^\dagger$:} Similar with SDEN, but the usage of $h$ is modified with Eq.\ref{equ:sden_h}. 


As we can see from Table \ref{table:results}, all contextual SLU models with memory mechanism can benefit from our dialogue logistic dependent multi-task framework, especially on the slot filling task. We also note that the improvement on intent detection is trivial, as single turn information has already trained satisfying intent classifiers according to results of NoMem in Table \ref{table:results}.
Thus, we mainly analyze DLI's impact on slot filling task and the prime metric is the F1 score.

In Table \ref{table:results}, the poorest contextual model is the SDEN, as its usage of the vector $h$ is too weak: simply initializes the BiLSTM tagger's hidden state with $h$, while other models concatenate $h$ with BiLSTM's input during each time step. The more the contextual model is dependent on $h$, the more obvious the improvement of the DLI task is. Comparing the performance of MemNet with $\text{SDEN}^\dagger$ on these two datasets, we can find that our $\text{SDEN}^\dagger$ is stronger than MemNet after the dialogue length increased.
Finally, we can see that improvements on KVRET* are higher than KVRET. This is because retrieving context knowledge from long-distance memory is challenging and our proposed DLI can help to consolidate the context memory and improve memory retrieval ability significantly in such a situation. 

We further analyze the training process of $\text{SDEN}^\dagger$ on KVRET* to figure out what happens to our model with DLI training, which is shown in Figure \ref{fig:loss on validation set}. We can see that the validation loss of $\text{SDEN}^\dagger+\text{DLI}$ falls quickly and its slot F1 score is relatively higher than the model without DLI training, indicating the potential of our proposed method.

To present the influence of hyper-parameter $\lambda$, we show SLU results with $\lambda $ ranging from 0.1 to 0.9 in Figure \ref{fig: lambda}. In this figure, we find that the improvements of our proposed method are relatively steady when $\lambda $ is less than 0.8, and 0.3 is the best one. When $\lambda $ is higher than 0.8, our model tends to pay much attention to the DLI task, overlook detail information within sentences, leading the SLU model to perform better on the intent detection but failing in slot filling.

\section{Conclusions}

In this work, we propose a novel dialogue logistic inference task for contextual SLU, with which memory encoding and retrieval components can be consolidated and further enhances the SLU model through multi-task learning. This DLI task needs no extra labeled data and consumes no extra inference time. Experiments on two datasets show that various contextual SLU model can benefit from our proposed method and improvements are quite impressive, especially on the slot filling task. Also, DLI is robust to different loss weight during multi-task training. In future work, we would like to explore more memory consolidation approaches for SLU and other memory related tasks.

\section*{Acknowledgements}

The research work descried in this paper has been supported by the National Key Research and Development Program of China under Grant No. 2017YFB1002103 and the Natural Science Foundation of China under Grant No. U1836221.

\bibliography{acl2019}
\bibliographystyle{acl_natbib}
\end{document}